# Generative Language Model for Catalyst Discovery

*Dong Hyeon Mok[1] and Seoin Back[1,*]*

[1]Department of Chemical and Biomolecular Engineering, Institute of Emergent Materials, Sogang University, Seoul 04107, Republic of Korea

Corresponding Author: sback@sogang.ac.kr (SB)

**Abstract**

Discovery of novel and promising materials is a critical challenge in the field of chemistry and material science, traditionally approached through methodologies ranging from trial-and-error to machine learning-driven inverse design. Recent studies suggest that transformer-based language models can be utilized as material generative models to expand chemical space and explore materials with desired properties. In this work, we introduce the Catalyst Generative Pretrained Transformer (CatGPT), trained to generate string representations of inorganic catalyst structures from a vast chemical space. CatGPT not only demonstrates high performance in generating valid and accurate catalyst structures but also serves as a foundation model for generating desired types of catalysts by fine-tuning with sparse and specified datasets. As an example, we fine-tuned the pretrained CatGPT using a binary alloy catalyst dataset designed for screening two-electron oxygen reduction reaction (2e-ORR) catalyst and generate catalyst structures specialized for 2e-ORR. Our work demonstrates the potential of language models as generative tools for catalyst discovery.

# 1. Introduction

The discovery of novel and advanced materials with superior performance surpassing that of existing materials is essential across various fields[1-4]. As new materials are newly developed, performance benchmarks rise, making it more challenging to identify better materials. This increases the difficulty and cost of traditional methods, such as experimental trial-and-error approach or computationally intensive simulations such as density functional theory (DFT) calculations[5].

In this context, machine learning (ML) has become a major technology for materials discovery, enabling more efficient exploration of vast chemical spaces. There are two primary ML-driven approaches to materials searches: high-throughput virtual screening (HTVS) and inverse design[6]. HTVS starts with a dataset within the screening scope and gradually narrows down the search space to focus on materials with desired properties. Initial candidates are filtered by multiple criteria, leaving only those that satisfy all requisite property thresholds as promising candidates. ML models enhance the screening process's efficiency by predicting which candidates are likely to fall outside the desired criteria, allowing for their early exclusion[7-10]. However, HTVS suffers from the drawback that the search scope is limited to the user-selected database. Expanding the chemical space through elemental substitution is a simple and common solution, yet this approach is unable to introduce structural diversity[9, 11].

Inverse design, in contrast, reverses the traditional material design process by starting with a desired set of properties. Unlike HTVS, which relies on a pre-defined set of candidates, inverse design leverages a broader search space and proposes new materials by refining structures until the desired properties are optimally achieved (global optimization), or by generating materials from a learned distribution of materials with desired properties (generative model)[12, 13]. In recent years, the rapid advancement of generative models, commonly used in creating images, audio, and video, has prompted significant interest in their applications within the materials domain[14-16]. Various generative techniques have been investigated, ranging from Variational Autoencoders (VAE)[17] and Generative Adversarial Network (GAN)[18] to diffusion models[19]. More recently, autoregressive language models based on Transformer architectures[20, 21], known for utilizing the self-attention mechanism and serving as the backbone of large language models (LLM), have shown promising results in generating both organic and inorganic materials[22, 23].

Since language models were originally designed to handle discrete sequences of variable lengths, they are particularly suited for generating crystal structures, which consist of discrete entities (atomic symbols) and can vary in length (number of atoms). This inherent flexibility makes language models more adept at this task compared to other generative models, such as diffusion models, which require additional considerations to handle the discrete and variable nature of crystal structures[24]. Flam-Shepherd and Aspuru-Guzik trained a GPT-based language model from scratch on sequences of tokenized atomistic data, including molecules, protein binding pockets and crystals[22]. They reported that their models performed comparably to or even surpassed other generative models. Antunes et al. demonstrated that CIF-format representation of crystals is also valid for training language models for generative purposes[25].

In this study, we developed a language model-based generative approach for catalyst discovery and demonstrated the potential of the language model as a foundation model for catalyst generation. We encoded catalyst structures into textual sequence data and trained a language model from scratch using a massive catalyst dataset. Since there is no standardized metric for evaluating catalyst generation performance, we developed an anomaly detection model to assess the validity of generated catalyst structures. To demonstrate the practical utility of our approach for catalyst discovery, we fine-tuned the pre-trained model using a small and biased dataset of binary alloy catalysts, specifically designed for two-electron oxygen reduction reaction (2e-ORR) catalysts[26, 27]. Finally, we evaluated the generated catalysts using machine learning potentials (MLP) and subsequent density functional theory (DFT) calculations, demonstrating that our strategy can generate new catalyst structures that have not been previously collected in database. This work highlights the potential of generative language models as useful tools for catalyst discovery.

## 2. Results and discussions

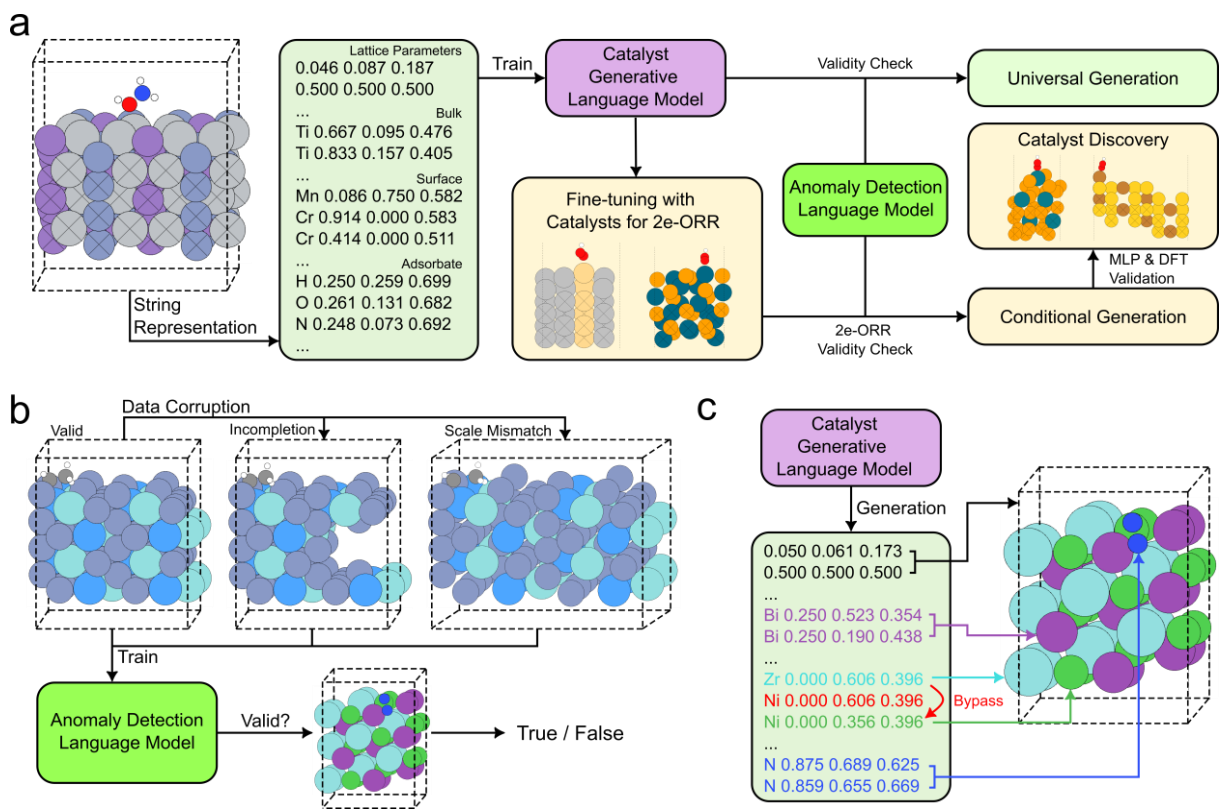

**Figure 1.** a) Schematic of the workflow for training a catalyst generative model based on a language model. The trained model was fine-tuned using a 2e-ORR dataset to apply the model for the specific catalyst discovery. After evaluating various validity metrics, candidate catalysts were validated with MLP predictions and DFT calculations. b) Schematic of the anomaly detection model to check the catalyst validity of generated structures. The detection model was trained on intentionally corrupted dataset to classify the validity of string representations of structures. c) Catalyst generation using a bypass method to address atomic overlapping. The generation of new atoms located close to already existing atoms was ignored in the process that converts string representations to 3D structures.

We implemented the Generative Pretrained Transformer 2 (GPT-2) architecture[28] to generate catalyst structures, hereafter referred to as 'CatGPT'. The CatGPT was trained autoregressively on two million catalyst structures containing both surface and adsorbate atoms, collected from the Open Catalyst 2020 Structure to Energy and Forces task dataset (OC20-S2EF 2M dataset)[29]. These structures were represented as a corpus consisting of tokenized strings of lattice parameters, atomic symbols and 3D coordinates (see **Methods** section). For validation, we used the Open Catalyst 2020 in-domain validation set (OC20-S2EF Val-ID dataset).

Since there are no systematically defined metrics for assessing the validity and quality of generated catalyst structures, we developed a model to detect invalidly generated structures based on Bidirectional Encoder Representations from Transformers (BERT)[30] and a validity metric for generated catalyst structures, hereafter referred to as the 'anomaly detection model' and 'catalyst validity', respectively. By fine-tuning the pre-trained CatGPT model using 2e-ORR dataset, we modified it to conditionally generate catalysts specifically for 2e-ORR, and evaluated the catalytic performance of the generated catalysts using MLP predictions and DFT calculations. The scheme of this work is summarized in **Figure 1a**.

**2.1. Evaluation metrics of generated structures**

Generative models for crystal structures have used two metrics to evaluate the validity of generated structures[31]. First, compositional validity assesses the charge neutrality of the system. Second, structural validity checks for overlapping atoms based on atomic distance criteria.

However, it should be noted that catalyst structures consisting of catalyst surface and adsorbate differ from bulk structures used in previous works. Generally, surface structures are prepared by cleaving bulk structures; thus, the charge neutrality of the system is not always guaranteed. In addition, as the CatGPT generates catalyst surfaces with adsorbates simultaneously, the compositional validity is not applicable. Although structural validity is partially useful, the following two invalid examples, identified in generated catalyst structures that passed the structural validity check, indicate that structural validity is necessary but not sufficient: (1) incomplete structures and (2) scale mismatches.

Incompletion occurs when the model terminates generation prematurely, resulting in structures with unusually empty regions or isolated atoms (Supplementary **Fig. 1a**). Scale mismatch refers to invalid generation when the lattice parameters are disproportionately large relative to the number of atoms in the unit cell, leading to disconnected atomic bonds (Supplementary **Fig. 1b**). Despite these anomalies being obviously recognized as unsuitable for catalyst structures by experts, there are no systematic criteria for identifying them.

Thus, we developed an anomaly detection model using BERT architecture to evaluate the catalyst validity of the generated structures (**Fig. 1b**). This model can detect incompletion and scale mismatch from string representation of catalyst structures. Because the model receives the string representation as input, it can directly evaluate the string representation of generated structures from the CatGPT without any conversion step. The model was trained to perform binary classification on a half-corrupted OC20 S2EF 2M dataset, where a random half of the original dataset was intentionally modified to represent anomalies of incompletion and scale mismatch. To create the incompletion feature, we randomly removed between 20 % and 80 % of atoms from the structures. For the scale mismatch feature, we expanded the unit cell dimensions by 150 % to 200 % while maintaining the fractional coordinates. The detection model demonstrated a high accuracy of 96 % in distinguishing valid and invalid catalyst structures for the half-corrupted validation set (Supplementary **Fig. 2a**). We employed this model to evaluate the catalyst validity of the generated structures. In addition, we found that some of the generated data were not successfully converted into 3D structures due to incorrect string representations. For example, some string representations had atomic symbols appearing twice sequentially, instead of the atomic symbol being followed by its coordinate sequence. Therefore, the ratio of valid atomic structures to generated structures is defined as 'generation validity'. Overall, we evaluated the validity of generated catalyst structures in three aspects: (1) generation validity, (2) structural validity and (3) catalyst validity.

Coverage and property distribution are key metrics for evaluating how well the generated data replicates the validation data, which serves as the ground-truth, meaning real catalyst structures. We used the method described by Xie et al.[31] to evaluate these metrics. Coverage assesses the structural and compositional similarity between the ensemble of generated and the ground truth structures, while property distribution evaluates the similarity in property distributions between generated and ground-truth structures. The structures and compositions are converted into vector forms using CrystalNN and Magpie fingerprints, respectively, as implemented in Matminer[32]. This allows us to calculate the similarity of catalysts arithmetically, enabling us to determine the coverage. Coverage is assessed by computing the pairwise distances between the vectorized structure and composition of the ground-truth structures and those of the generated structures[33]. Coverage Recall (Precision) quantifies the proportion of ground-truth (generated) structures whose feature vector distances to the nearest generated (ground-truth) structures fall within a predefined cutoff criterion (0.2 for structure and 8.5 for composition, respectively). Property distribution is evaluated by computing the Earth Mover's Distance (EMD)[34] between the property distributions of the generated and ground-truth structures sets. Density ($\rho$, g/cm$^3$) and number of unique elements ($N_{el}$) of structures are used as properties.

To establish the upper and lower bounds of the metrics and to rationalize the anomaly detection model as a sanity check for catalyst validity, we computed these metrics on the validation set. The catalyst validity score of 0.997 observed for the validation set supports the suitability of this metric for evaluating the quality of generated catalyst structures (**Table 1**). Notably, the low compositional validity score for validation structures (0.133) confirms that the charge neutrality is not a reliable measure for assessing the valid generation of catalyst structures.

## 2.2. Performance of generations by CatGPT

We generated 10,000 catalyst structures using the '<bos>' token, which indicates the beginning of catalyst sequences, and evaluated their validity, coverage and property distributions. Validity and coverage were evaluated for the structures that satisfied the generation validity criteria, while property distributions were assessed for 5,000 valid structures randomly sampled from those that satisfied all validity criteria.

The results demonstrated that CatGPT performs exceptionally well in generating high-quality catalyst structures and accurately retrieving a distribution similar to the validation set as evidenced by nearly perfect precision and recall scores, respectively. However, we observed that it occasionally generates catalyst structures with overlapping atoms, resulting in a low structural validity score (**Table 1**). To address the generation of overlapping atoms without modifying the model architecture, we introduced a new feature to bypass overlapping atoms when converting string representations of catalysts into 3D structures (**Fig. 1c**). This feature takes advantage of the autoregressive nature of the language model, which predicts tokens sequentially, unlike other generation models that create entire structures at once by decoding latent vectors or denoising Gaussian noise. The generative model with this feature, named CatGPT-BP, shows perfect structural validity with only a minimal decrease in the catalyst validity score. Additionally, the slightly lower catalyst validity value compared to the validation set, despite nearly maximum structural validity and coverage, highlights the difficulty of detecting anomalies using other metrics and the necessity of the catalyst validity metric.

**Table 1.** Catalyst structure generation performance of CatGPT and CatGPT-BP. To calculate coverage and property distributions of the generated structures, 10,000 randomly sampled structures from the validation set (OC20 S2EF Val-ID dataset) were used as the ground truth. The coverage and property distribution of validation set were calculated with respect to 10,000 randomly sampled structures from the training set.

| Method | Validity of Generated Structures | | | Coverage | | Property Distribution | |
|---|---|---|---|---|---|---|---|
| | Generation Validity↑ | Structural Validity↑ | Catalyst Validity↑ | Recall↑ | Precision↑ | EMD ($\rho$)↓ | EMD ($N_{el}$)↓ |
| (Validation set) | - | 1.000 | 0.997 | 0.999 | 1.000 | 0.014 | 0.019 |
| CatGPT | 0.997 | 0.686 | 0.917 | 0.999 | 0.999 | 0.302 | 0.028 |
| CatGPT-BP | 0.997 | 1.000 | 0.906 | 0.999 | 0.998 | 0.435 | 0.046 |

Since an ideal generative model should be invariant to translation (or rotation) and permutation, we prepared two augmented datasets to introduce translational and permutational features (details in the **Methods** section). As shown in the training loss curves (Supplementary **Fig. 3a**), while the translation and rotation of structures barely affected performance, permutational augmentation, which involves shuffling the order of atoms, clearly decreased the performance. The difficulty in accounting for permutation invariance in CatGPT implies that the sequence in the string representation of catalyst structures is itself informative.[24] This is more clearly demonstrated in the generation probability distribution of the end token (Supplementary **Fig. 3b**).

In the model trained without the augmented permutation dataset, the tokens of adsorbate atoms increased the probability of end token generation, effectively acting as indicators for the termination of generation. However, in the model trained with the augmented permutation dataset, the probability increased gradually with sequence length, making it difficult to identify the end of generation. This led to repeated generations of atoms in the same position. As a result, the model augmented with the permutation dataset demonstrated relatively poor generative performance (Supplementary **Table 1**).

## 2.3. Fine-tuning CatGPT-BP for discovery of 2e-ORR catalysts

In the field of materials inverse design via generative models, it is considered more practical to focus on generating catalysts for specific purposes rather than covering the entire chemical space[6]. With this perspective, we aimed to fine-tune the model to understand the rules within a small and biased dataset and generate desired catalyst structures with reasonable validity and diversity. In our previous works, it was proposed that two-electron oxygen reduction reaction (2e-ORR), which electrochemically produces $H_2O_2$ from $O_2$, is facilitated on single-site binary alloys consisting of oxophilic and oxophobic elements[26, 27]. The ensemble effect of isolated oxophilic sites weakens the Gibbs free energy of O* ($\Delta G_{O*}$), a selectivity descriptor of 2e-ORR, by enforcing the ontop site adsorption of O*. When $\Delta G_{O*}$ exceeds $\Delta G_{H2O2*}$ (3.52 eV), maximum selectivity is considered achieved. The ligand effect of oxophobic elements surrounding the active site controls the Gibbs free energy of OOH* ($\Delta G_{OOH*}$), an activity descriptor of 2e-ORR, which exhibits maximum activity when its value is 4.22 eV.

To investigate these effects and discover promising 2e-ORR catalysts, Back et al.[26] previously generated a binary alloy dataset based on two rules: (1) the binary alloys should consist of oxophobic and oxophilic elements (composition rule) and (2) the active sites should be the ontop site of oxophilic atoms (adsorption rule). To ensure that the learned distribution of the fine-tuned model follows the fine-tuning dataset distribution, we defined composition and adsorption validity as additional metrics to confirm whether the generated catalyst structures satisfy these rules. We note that the catalyst anomaly detection model to evaluate the catalyst validity was also fine-tuned with a half-corrupted 2e-ORR dataset and achieved high accuracy (Supplementary **Fig. 2b**).

The results from 1,000 sampled structures demonstrated that fine-tuning the pre-trained model can modify the generation data distribution toward the fine-tuning data while maintaining or outperforming the generation performance of the pre-trained model (**Table 2**). Particularly, a composition and adsorption validity score of over 0.95 indicates that the model successfully learns the rules of the fine-tuning dataset and generates appropriate catalyst structures for 2e-ORR with only about 1,700 fine-tuning data points. Additionally, it was confirmed that the bypassing method remains a practical approach for the fine-tuned model.

**Table 2.** Evaluated metrics of fine-tuned CatGPT models using the 2e-ORR dataset. As in **Table 1**, the structural, catalyst, 2e-ORR validity metrics and coverages were evaluated for the sampled structures satisfying the generation validity, while property distribution were assessed for 500 valid structures randomly sampled from those satisfying structural and catalyst validity. The coverage and property distribution of the validation set were calculated with respect to 1,000 randomly sampled structures from the training set.

| Method | Validity | | | Coverage | | Property Distribution | | 2e-ORR Validity | |
|---|---|---|---|---|---|---|---|---|---|
| | Generation↑ | Structural↑ | Catalyst↑ | Recall↑ | Precision↑ | wdist($\rho$)↓ | wdist($\underline{N}_{el}$)↓ | Composition | Adsorption |
| (Validation set) | - | 1.000 | 0.988 | 1.000 | 1.000 | 0.111 | 0.000 | 1.000 | 1.000 |
| CatGPT | 0.986 | 0.980 | 0.976 | 0.971 | 0.982 | 0.302 | 0.000 | 0.996 | 0.975 |
| CatGPT-BP | 0.986 | 1.000 | 0.974 | 0.971 | 0.981 | 0.259 | 0.000 | 0.996 | 0.974 |

## 2.4. Generation strategy

Considering that the purpose of generative models is to discover novel catalysts, the diversity, uniqueness and novelty of generated catalysts are as important as valid generation. Since the probability distribution of autoregressive models can be modified by generation parameters, such as temperature and input queries, we compared the validity and diversity of generated structures from the 2e-ORR fine-tuned model under different temperatures and queries. To evaluate diversity, we used metrics of uniqueness and novelty.[35] Uniqueness is the proportion of validly generated structures that are unique ($N_{unique}/N_{valid}$), while novelty is the proportion of validly generated structures that are not contained in the training set ($\underline{N}_{novel}/N_{valid}$). The detection of duplicated structures in both cases was performed by 'StructureMatcher' module implemented from Pymatgen[36].

To observe the effect of input queries, we collected the lattice parameters of 1,000 structures randomly selected from the OC20 S2EF Val-ID set, and used them as queries for generating new structures. Note that these lattice parameters were not included in the fine-tuning dataset. Because the conditional probabilities of generation vary with the given lattice parameters, the uniqueness and novelty of the model greatly increased. However, the proportion of invalid structures also increased, with around 60 % of structures being classified as anomalous based on catalytic validity metric (**Table 3**). This result indicates that the fine-tuned model is weak in extrapolating validly generated out-of-domain catalyst structures whose lattice parameters are not covered in the fine-tuning dataset distribution.

**Table 3.** Validity, coverage and diversity of the fine-tuned CatGPT models at different temperatures and input queries. Because the metrics were assessed using the bypassing method, the structural validities are 1.00, thus they are omitted from the table. The metrics assessed without the bypassing method can be found in supplementary **Table 2.** The diversity metrics were assessed for the samples that satisfy all validity criteria. 'Lattice' indicates the result of generated structures sampled from lattice parameters, while the others were sampled from '<bos>' tokens.

| Temperature | Validity | | 2e-ORR Validity | | Coverage | | Diversity | |
|---|---|---|---|---|---|---|---|---|
| | Generation↑ | Catalytic↑ | Composition↑ | Adsorption↑ | Recall↑ | Precision↑ | Uniqueness↑ | Novelty↑ |
| 1.0 | 0.986 | 0.974 | 0.996 | 0.974 | 1.000 | 0.997 | 0.208 | 0.091 |
| 1.0 (Lattice) | 0.965 | 0.416 | 0.807 | 0.626 | 0.988 | 0.892 | 0.991 | 0.926 |
| 0.5 | 1.000 | 0.991 | 1.000 | 0.990 | 0.977 | 1.000 | 0.054 | 0.108 |
| 0.7 | 0.999 | 0.990 | 0.999 | 0.988 | 0.994 | 0.999 | 0.100 | 0.083 |
| 1.5 | 0.962 | 0.959 | 0.995 | 0.919 | 1.000 | 0.993 | 0.403 | 0.206 |
| 2.0 | 0.943 | 0.844 | 0.938 | 0.800 | 0.994 | 0.980 | 0.733 | 0.492 |

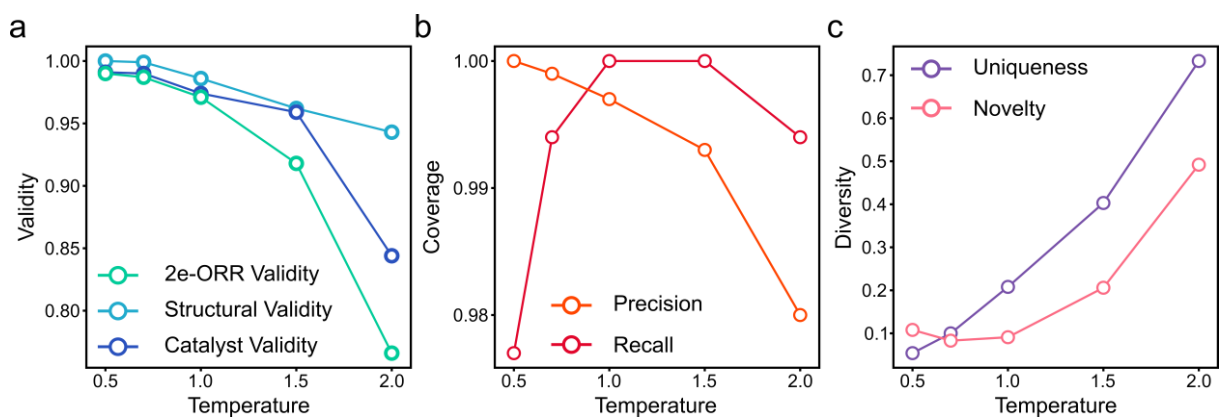

**Figure 2.** a) Validity, b) coverage and c) diversity as a function of sampling temperatures. 2e-ORR validity is the ratio of generated structures satisfying both adsorption and composition validity.

The effect of generation parameters is measured by changing the temperatures (τ), which control the confidence of the model by adjusting the probability distribution of the next token prediction within the softmax function. At low temperatures, the probability distribution of the next token prediction becomes sharper, resulting in more valid generations but with less diversity. In contrast, high temperatures flatten the distribution, leading to higher diversity of predictions but less valid generations. We confirmed this trade-off between diversity and validity from the results of the fine-tuned model, both with and without the bypassing method

(**Fig. 2** and Supplementary **Table 2**). Additionally, because temperature modifies the probability distribution of the trained model, the change in temperature affects the coverage. Coverage Recall decreased at both low and high temperatures, either because the generated structure distribution is too localized or too broad, respectively, to recover the validation data. Coverage Precision, which indicates the ratio of high-quality structures, follows the same trend as validity.

### 2.5. Application of CatGPT-BP for discovery of 2e-ORR catalysts

We validated the generated catalyst structures by performing geometry optimizations and calculating $\Delta G_{OOH*}$ and $\Delta G_{O*}$. Instead of directly performing computationally expensive density functional theory (DFT) calculations, we initially employed machine learning potential (MLP), specifically the pre-trained EquiformerV2 model[37]. Based on the MLP-optimized structures and the predicted binding Gibbs free energies ($\Delta G_{OOH*}$ and $\Delta G_{O*}$), we collected catalysts satisfying the activity condition with a considerable margin (3.22 eV < $\Delta G_{OOH*, MLP}$ < 5.22 eV), and subsequently performed DFT calculations to obtain both $\Delta G_{OOH*}$ and $\Delta G_{O*}$.

From the 1,000 generated catalyst structures with a temperature parameter of 1.5, 968 structures were successfully converted into atomic structures in 3D space, and 858 structures satisfied all validity metrics. Among these valid structures, we identified 133 unique and novel structures. MLP optimization and binding energy prediction resulted in 35 structures satisfying the activity criteria. Subsequently, clean catalyst surfaces and O* adsorbed surfaces were constructed, and DFT optimizations were performed to calculate $\Delta G_{OOH*}$ and $\Delta G_{O*}$. Among them, 10 catalyst structures were successfully optimized within a time constraint (24 hours or 999 relaxation steps) (**Fig. 3a** and **b**). DFT results confirmed that all these structures met the activity and selectivity criteria, with five exhibiting significant 2e-ORR activity in the range of 4.22 ± 0.2 eV (**Fig. 3c**). We note that none of these structures exist in the training data, and the suggested binary alloy combinations have not been reported as 2e-ORR catalysts in literature. This result highlights that the language model-based catalyst generative model is a potential platform for the discovery of novel catalysts via fine-tuning with a relatively small dataset.

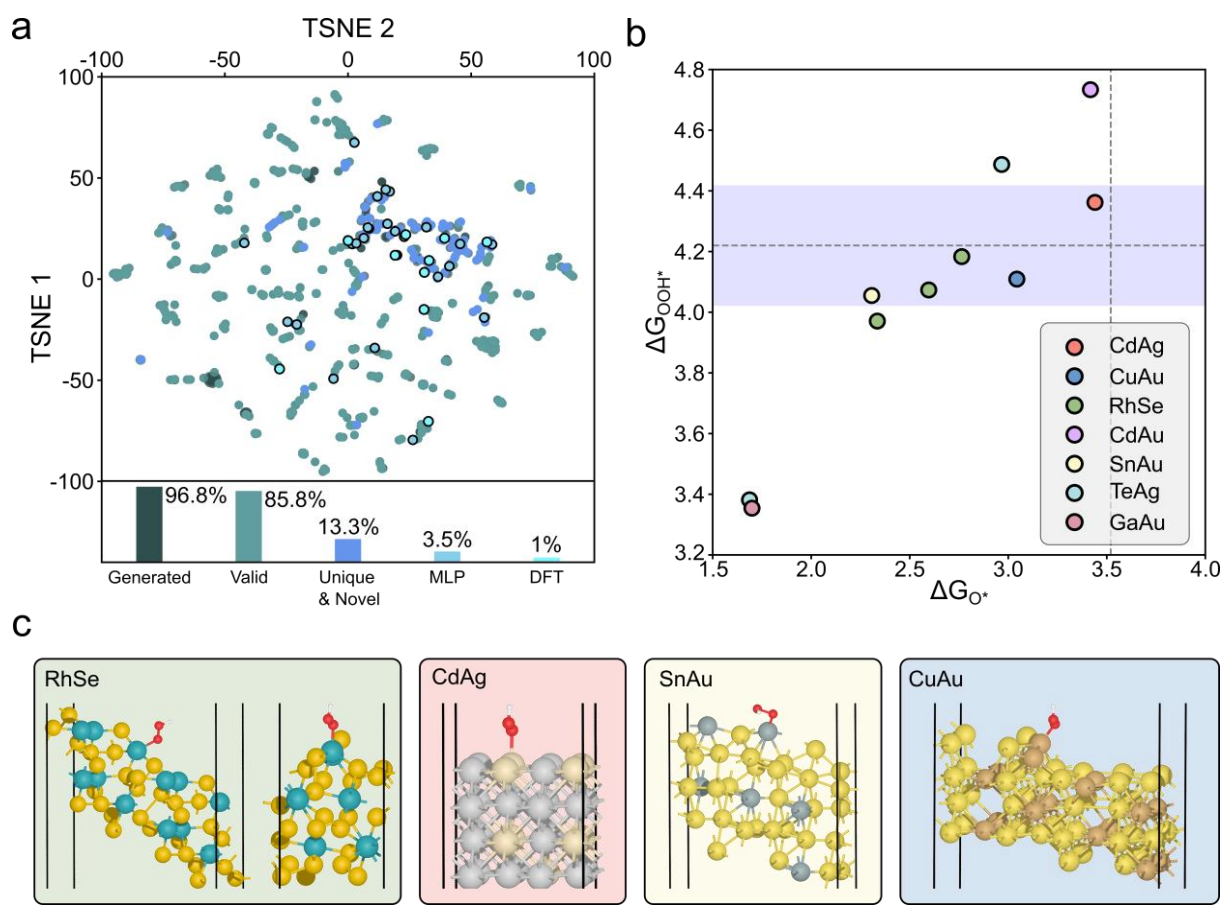

**Figure 3.** a) Two-dimensional visualization of the screening process using structure featurization and dimension reduction with t-distributed stochastic neighbor embedding (t-SNE)[38]. The bar graph below shows the percentage of the remaining structures in each screening scheme. b) DFT calculated $\Delta G_{O*}$ and $\Delta G_{OOH*}$ of 10 catalysts. The horizontal dashed line indicates $\Delta G_{OOH*}$ to achieve the maximum 2e-ORR activity, while the vertical dashed line indicates $\Delta G_{O*}$ to achieve the maximum 2e-ORR selectivity. c) The atomic structures of active catalysts whose activities are close to the maximum (4.22 ± 0.2 eV, visualized as the blue area in **Figure 3b**).

## 2.6. Challenges and future directions

While catalysts can be easily modeled using generated crystal structures from crystal generative models, it is very challenging to recover bulk crystals from generated catalyst structures. This makes it infeasible to obtain bulk information and evaluate bulk properties such as space group and formation energy, which are important for proposing novel catalysts, requiring a development of new models to achieve these goals. Furthermore, although the

CatGPT generates new structures, it struggles to suggest new combinations outside of the training data, restricting the model's role in expanding elemental diversity.

We noticed that a prompt engineering approach can enhance the functional flexibility of LLMs.[39] For example, Gruver et al. showed that elemental substitution via the infilling task of fine-tuned LLMs resulted in a higher percentage of materials with desired properties compared to random substitution, indicating the potential for conditional generation with structural and compositional chemical space expansion.[24] Other strategies to leverage LLMs for inorganic materials can also be applied to catalyst systems.[40, 41]

3. Conclusions

While several works have progressed in developing generative language models for materials discovery, none have applied these models to catalyst surface systems. This may be due to a lack of validation methods for complex systems, which have more atoms and greater structural diversity than simple crystals, and substantial data requirements (at least 10,000 samples) for training and validating generative models on their systems.

In this work, we introduce the autoregressive language models for generating catalyst structures. The results demonstrated that the catalyst generative model based on GPT-2 could generate catalysts close to real structures but struggled to generate valid structures without atomic overlapping. We addressed the overlapping issue with the bypassing method that leverages the sequential nature of the language model without degrading its overall generative performance. Furthermore, we showed that the model can be utilized as a universal catalyst generative pre-trained model for a fine-tuning strategy. With only about 2,000 fine-tuning data points of 2e-ORR catalyst structures, the fine-tuned model successfully generated structures within the fine-tuning data distribution while maintaining the generative performance, suggesting novel and promising catalysts for 2e-ORR. These results suggest the potential of autoregressive language models not only as generative model for catalyst structures but also as foundation model for catalyst discovery.

## 4. Methods

**Representations of Catalyst Structures**

Bulk crystal structures can be represented as three-dimensionally repeating systems[31]. Similar to crystals, catalyst surface structures can be fully represented by the lattice parameters of the unit cell, atomic symbols and respective fractional (or absolute) coordinates of atoms. Thus, the tuple C, for representing catalyst surfaces consisting of N atoms, is configured as follows:

$C = (l_1, l_2, l_3, \theta_1, \theta_2, \theta_3, e_1, x_1, y_1, z_1, ..., e_N, x_N, y_N, z_N)$,

This includes the length ($l_1$, $l_2$, $l_3$) and angle ($\theta_1$, $\theta_2$, $\theta_3$) information of the unit cell lattice parameters, atomic symbols ($e_i$) and their fractional coordinates set ($x_i$, $y_i$, $z_i$) in the unit cell.

During the tokenization process, the components in tuple $C$ are represented as string tokens within the vocabulary space, which comprises textual tokens for all atomic symbols in the periodic table, numerical tokens for the lattice parameters and coordinates of the crystal. Among the various tokenization strategies for numbers, we employed coordinate-level tokenization, which directly represents each fraction as a distinct token, for example, '0.000', 0.001', ..., '1.000'. To apply these tokenization strategies to lattice parameters (lengths and angles), we divide them by 180 so that they share the same numerical space as the fractional coordinates. Since the ordering of atoms itself can hold structural information in the catalyst structures, the pairs of atomic symbols and coordinates in tuple C are arranged in the order of bulk, surface, and adsorbate. We note that atoms in the first layer, excluding adsorbates, were categorized as surface atoms, while the rest were treated as bulk atoms, following the method of Ref[29].

**Model Structures and Training**

We used the GPT-2 architecture[28] from HuggingFace to generate 3D catalyst structures tokenized as components of tuple C. The tokens $t_i$ in the sequence of tokenized structures are predicted by a categorical probability distribution $p(t_i|t_{0:i-1})$, where $t_{0:i-1}$ is the sequence of input tokens up to $i-1$. To generate a sequence of tokenized structure $x$, the joint probability is modeled by the transformer as follows:

$$p(x) = \prod_{i=1}^{n} p(t_i|t_{0:i-1}) \tag{1}$$

Task conditioning of GPT-2 for multitask learning was not utilized in this work, so the only difference from GPT-1[20] model is the architecture, such as the position of layer normalization in each transformer decoder block.

For the anomaly detection model, we utilized BERT. This architecture is designed to predict masked tokens and thereby understand the inherent features of data, making it suitable for classification tasks rather than generation. To detect anomalies of incompletion and scale mismatch, we added a regression head to the embedding of the initial token, enabling the detection of anomalies from the generated string representations of catalyst structures.

We trained both GPT-based generative model and BERT-based detection model with 12 self-attention layers, 8 attention heads, an embedding size of 512 and a batch size of 144 on Geforce RTX 3090 GPU. Training epochs are set to 30 and 10, respectively. The other parameters were set to their default values in HuggingFace.

**Dataset Details**

To train the scratch GPT-2 model to generate string representations of catalyst structures, we utilized OC20-S2EF dataset that contains single-point calculated heterogenous catalyst structures generated during the DFT relaxation[29]. Although there are about 250 million data points in the OC20 database, we used 2 million catalyst structures to train the CatGPT considering the training cost (1.48 hr/1K batch steps on one RTX 3090 GPU). For validation, OC20-S2EF Val-ID dataset was used.

For training the BERT as an anomaly detection model, 50 % of the OC20-S2EF 2M dataset was randomly sampled and corrupted in two ways (incomplete structures and scale mismatches) in equal proportions. These corrupted samples were labeled as 0, and the other 50 % was left unchanged and labeled as 1. The same scheme was applied to the OC20-S2EF Val-ID dataset for validation of the detection model.

To fine-tune the GPT-2 model pre-trained with OC20-S2EF 2M dataset, we used a dataset of 1,721 catalyst structures generated for screening two-electron oxygen reduction reaction catalysts in our previous studies (2e-ORR dataset)[26]. The 2e-ORR dataset consists of binary alloy surface structures composed of oxophilic and oxophobic elements, featuring oxophilic active sites surrounded by oxophobic atoms (Supplementary **Fig. 4**). We randomly selected 80 % of the 2e-ORR dataset as the training set and 10 % as the validation set to fine-tune the pre-trained model. The remaining 10 % was used as the test set to evaluate the performance of the fine-tuned model. The same corruption scheme described above was applied to the 2e-ORR dataset for fine-tuning the anomaly detection model.

**Metrics Details**

Structural validity checks the overlapping of atoms based on atomic distances and cell volume. The cutoff values for atomic distances and cell volume to determine atomic overlap are set to 0.5 Å and 0.1 Å$^3$, respectively. Catalyst validity is determined by the anomaly detection model, which classifies catalyst structures to be anomalous if the predicted values of catalyst structures are below 0.5. Coverage measures the similarity between generated and ground-truth structures using the pairwise distance of vectorized structures and compositions by CrystalNN and Magpie fingerprints, respectively. The CrystalNN fingerprint converts periodic structures into vectors that represent the local coordination information of sites within the structures[42]. The Magpie fingerprint converts the element stoichiometry of structures into weighted elemental properties based on stoichiometric attributions[43]. The distance cutoffs were set to 0.2 for structure and 8.5 for composition, which correspond to about 2/3 of the average pairwise distances of structure and composition vectors (0.380 and 13.356, respectively) of 10,000 randomly sampled OC20-S2EF 2M dataset.

**Data Augmentations**

To ensure that generative models for catalyst structures maintain invariance to both unit cell rotation and spatial translation of atomic coordinates within lattice boundaries (rotation and translation invariance, respectively), we constructed an augmented training and validation dataset. Specifically, 33% of the augmented dataset undergoes a uniform translation of atoms in randomly selected vector directions, while another 33% undergoes rotations within the x-y plane of the unit cell at random scales. The remaining dataset was left unchanged.

Ideally, the order of atoms should not affect the generative performance of the model (permutation invariance). However, due to the nature of autoregressive language models, which generate atoms sequentially, we discovered that atom ordering significantly influences the generation outcomes. To investigate the impact of atom sequencing on generative performance, we prepared a dataset wherein the atomic orderings were randomly shuffled throughout the entire dataset.

**Computational Details**

To evaluate the 2e-ORR activity and selectivity of screened structures, spin-polarized DFT calculations were performed using the Vienna Ab initio Simulation Package (VASP, version 5.4.4)[44, 45] with the projected-augmented wave (PAW) pseudopotential method[46] and the generalized gradient approximation revised Perdew-Burke-Ernzerhof (GGA-RPBE) exchange-correlation functional[47]. Geometry optimizations were conducted with convergence criteria of $10^{-4}$ eV for energy and 0.05 eV/Å for force, and a kinetic energy cutoff of 400 eV was applied. Monkhorst-Pack k-point mesh[48] was configured ($k_1 \times k_2 \times 1$) to ensure 25 Å < $a_n \times k_n$ (n = 1, 2) < 30 Å, where $a_1$ and $a_2$ represent the unit vector sizes in the $x$ and $y$ directions, respectively.

To calculate the Gibbs free energy of adsorbates, the computational hydrogen electrode (CHE) method[49] was used, which equates the chemical potential of ½H$_2$ (g) with (H$^+$ + e$^-$) at 0 V$_{RHE}$ under standard conditions. The Gibbs free energy of O* and OOH* adsorption was calculated as follows:

$$\Delta G_{O^*} = E_{O^*} - E_{slab} - E_{H_2O} - E_{H_2} + \Delta(ZPE + \int C_p dT - TS)$$

$$\Delta G_{OOH^*} = E_{OOH^*} - E_{slab} + 2E_{H_2O} - \frac{3}{2}E_{H_2} + \Delta(ZPE + \int C_p dT - TS)$$

where $E_{O^*}$ and $E_{OOH^*}$ are DFT energies of O* and OOH* adsorbed surfaces, and $E_{H_2}$ and $E_{H_2O}$ are DFT energies of H$_2$ and H$_2$O molecules, respectively. Zero-point energies (ZPE), enthalpic contribution ($\int C_p dT$), and entropic contribution (TS) were calculated using the Harmonic oscillator and Ideal gas approximations for adsorbates (O*, OOH*) and gas molecules (H$_2$, H$_2$O), respectively, as implemented in Atomic Simulation Environment (ASE)[50]. The correction values are given in Supplementary **Table 3**. In addition, since the MLP model used in this work predicts the DFT binding energy ($\Delta E_{OOH^*,ML}$) directly, the ML predicted Gibbs free energy of OOH* adsorption was calculated as follow:

$$\Delta G_{OOH^*,ML} = \Delta E_{OOH^*,ML} + \Delta(ZPE + \int C_p dT - TS)$$

A solvation effect for OOH* (−0.25 eV) was also included[51]. Due to the poor description of the DFT energy of O$_2$ by GGA functionals[52], an experimental reaction Gibbs free energy of ORR was used instead, with $\Delta G_{rxn}$ values of −4.92 eV for O$_2$ + 2H$_2$ → 2H$_2$O and

−1.40 eV for $O_2 + H_2 \rightarrow H_2O_2$.

## 5. Data Availability

OC20 dataset can be found in https://fair-chem.github.io/core/datasets/oc20.html. 2e-ORR dataset can be found in https://github.com/SeoinBack/CatGPT

## 6. Code Availability

The code developed in this work and relevant information can be found in Github (https://github.com/SeoinBack/CatGPT).

## 7. Acknowledgements

S.B. acknowledges the support from the Nano & Material Technology Development Program through the National Research Foundation of Korea (NRF) funded by Ministry of Science and ICT (RS-2024-00406517).